\pdfoutput=1

\documentclass[11pt]{article}

\usepackage{EMNLP2022}

\usepackage{times}
\usepackage{latexsym}
\usepackage{amsfonts}
\usepackage{graphicx}
\usepackage{url} 
\usepackage{amsmath}
\usepackage{ulem}
\usepackage{booktabs}
\usepackage{multirow}
\usepackage{enumitem}
\usepackage{marvosym}
\usepackage{microtype}

\usepackage{listings}

\usepackage{xcolor}

\definecolor{codered}{rgb}{1,0,0}
\definecolor{codegreen}{rgb}{0,0.6,0}
\definecolor{codegray}{rgb}{0.5,0.5,0.5}
\definecolor{codepurple}{rgb}{0.58,0,0.82}
\definecolor{backcolour}{rgb}{0.95,0.95,0.92}

\lstdefinestyle{mystyle}{
 backgroundcolor=\color{backcolour}, commentstyle=\color{codegreen},
 keywordstyle=\color{magenta},
 numberstyle=\tiny\color{codegray},
 stringstyle=\color{codepurple},
 basicstyle=\ttfamily\footnotesize,
 breakatwhitespace=false,     
 breaklines=true,         
 captionpos=b,          
 keepspaces=true,         
 numbers=left,          
 numbersep=5pt,         
 showspaces=false,        
 showstringspaces=false,
 showtabs=false,         
 tabsize=2
}
\title{DistilCSE: Effective Knowledge Distillation For Contrastive Sentence Embeddings}

\author{
   Chaochen Gao\textsuperscript{\rm 1,2}, Xing Wu\textsuperscript{\rm 1,2,4}, Peng Wang\textsuperscript{\rm 1,2}\thanks{The first three authors contribute equally.} \\
   \textbf{Jue Wang}\textsuperscript{\rm 3}, \textbf{Liangjun Zang}\textsuperscript{\rm 1}, \textbf{Zhongyuan Wang}\textsuperscript{\rm 4}, \textbf{Songlin Hu}\textsuperscript{\rm 1,2 \Letter}
  \\
  \textsuperscript{\rm 1}Institute of Information Engineering, Chinese Academy of Sciences\\
  \textsuperscript{\rm 2}School of Cyber Security, University of Chinese Academy of Sciences\\
  \textsuperscript{\rm 3}Zhongnan University of Economics and Law, 
  \textsuperscript{\rm 4}Kuaishou Technology\\
  \{gaochaochen,wuxing,wangpeng,zangliangjun,husonglin\}@iie.ac.cn \\
    201821090281@stu.zuel.edu.cn, wangzhongyuan@kuaishou.com
}

\begin{document}
\maketitle
\begin{abstract}
Large-scale contrastive learning models can learn very informative sentence embeddings, but are hard to serve online due to the huge model size.
Therefore, they often play the role of ``teacher", transferring abilities to small ``student" models through knowledge distillation.
However, knowledge distillation inevitably brings some drop in embedding effect.
To tackle that, we propose an effective knowledge distillation framework for contrastive sentence embeddings, termed \textbf{DistilCSE}. It first applies knowledge distillation on a large amount of unlabeled data, and then fine-tunes student models through contrastive learning on limited labeled data. 
To achieve better distillation results, we further propose Contrastive Knowledge Distillation (\textbf{CKD}). CKD uses InfoNCE as the loss function in knowledge distillation, enhancing the objective consistency among teacher model training, knowledge distillation, and student model fine-tuning. 
Extensive experiments show that student models trained with the proposed DistilCSE and CKD suffer from little or even no performance decrease and consistently outperform the corresponding counterparts of the same parameter size. Impressively, our 110M student model outperforms the latest state-of-the-art model, i.e., Sentence-T5 (11B), with only 1\% parameters and 0.25\% unlabeled data.
\end{abstract}

\section{Introduction}
Sentence embeddings trained with contrastive learning provide dense vector representations widely applied in many real-world applications (like sentence textual similarity, text retrieval, etc.). State-of-the-art (SOTA) contrastive sentence modeling methods \cite{gao2021simcse, wang2021aligning, yan2021consert} that achieve remarkable performance are all based on Pretrained Language Models (PLMs)\cite{devlin2018bert,liu2019roberta,raffel2019exploring}. Moreover, PLMs-based sentence embedding methods tend to use larger model sizes and training data scales for better performance. For instance, the latest SOTA model, i.e., Sentence-T5 \cite{ni2021sentence}, is built with 11 billion parameters and trained on 2 billion question-answer pairs. Though effective, such large models are hard to be applied in real-world applications with limited computational resources or time cost for model inference. Therefore, they often play the role of ``teacher", transferring their high-capacities to small ``student" models through knowledge distillation (KD) \cite{romero2014fitnets, kim2016sequence, hu2018attention, sanh2019distilbert, sun2020mobilebert, wang2020minilm, jiao2019tinybert}. 
Typically, KD is conducted on the same training data as the teacher model is built on \cite{sun2020mobilebert}, and the student models are trained to mimic the behaviour of the teacher models in the output projection layers or in the intermediate hidden layers.

With the development of KD methods, student models perform very close to or even exceed teacher models on many tasks \cite{jiao2019tinybert, sun2020contrastive, pan2020meta}. But these works are mainly focused on classification tasks, and there are few KD optimizations for contrastive sentence embeddings\footnote{Contrastive learning has emerged as a promising trend in learning sentence embeddings and contrastive sentence embedding \cite{gao2021simcse} has achieved very good results.}.
Therefore, we explore the effect of current state-of-the-art KD methods\cite{hinton2015distilling, sun2019patient,pan2020meta} on the contrastive sentence modelling task, the experimental results can refer to Appendix A. Though these methods achieve very good results on other tasks, but they inevitably bring some drops in embedding effect. So it is challenging to adequately transfer the capability of a large contrastive sentence embedding model to a small student, especially on the limited labeled data used for training the teacher model.

\begin{figure*}[!htbp]
\centering
\includegraphics[width=0.7\textwidth,height=0.28\textheight]{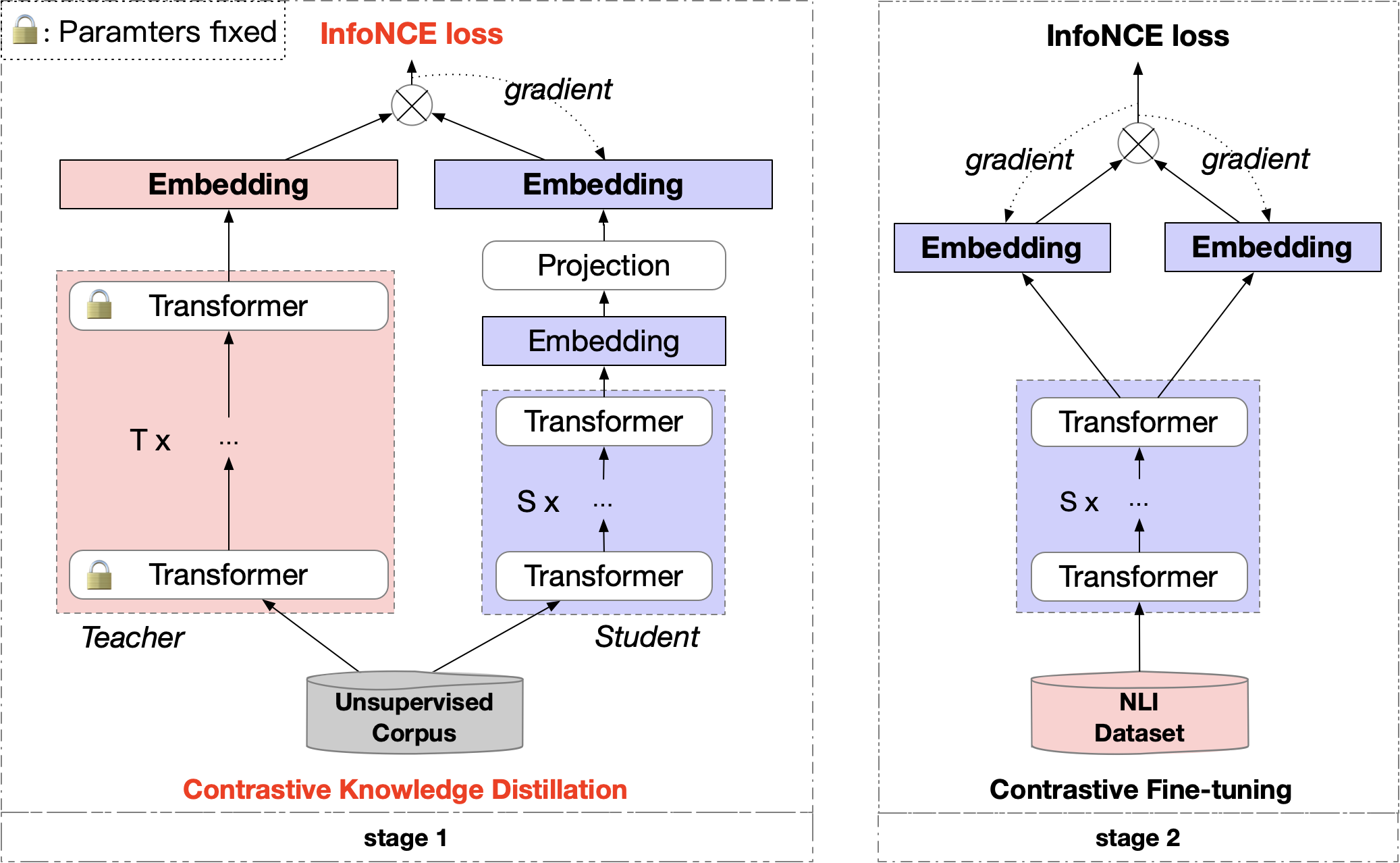}
\caption{The proposed DistilCSE framwork and Contrastive Knowledge Distillation (CKD) method. Different from the previous practice of directly distilling on labeled NLI dataset, DistilCSE consists of two stages: knowledge distillation on large unsupervised corpus (stage 1) and further finetuning on NLI dataset with contrastive learning (stage 2). Unlike previous distillation methods, the CKD method uses the InfoNCE loss function instead of the MSE loss function to enhance the objective consistency among teacher model training, knowledge distillation, and student model fine-tuning.}
\label{DistilCSE}
\end{figure*}

To alleviate the performance drop, we propose \textbf{DistilCSE}, a simple but effective knowledge \underline{distil}lation framework for \underline{c}ontrastive
\underline{s}entence \underline{e}mbedding.
As shown in Figure \ref{DistilCSE}, the proposed DistilCSE framework consists of two stages. At stage 1, we conduct KD to transfer the capability of a well-trained large contrastive sentence embedding model to a small student model, using a large set of unlabeled data for adequate knowledge transfer. 
At stage 2, the student model is further fine-tuned with supervised contrastive learning on labeled data to alleviate the domain bias brought by the unlabeled data. Specifically, labeled data refers to the training data of the teacher model.

Further inspired by the performance improvement brought by the consistency of training objectives between prompt learning and autoregressive PLMs (e.g., GPT-3), we propose a novel KD method termed Contrastive Knowledge Distillation (CKD). 
As illustrated in the Figure \ref{DistilCSE}-(b), the proposed CKD uses an identical loss function of contrastive learning, i.e., the InfoNCE loss \cite{hjelm2018learning}, for the distillation process.
Such a setting brings consistency of the objective functions in different processes:  the teacher model's training, the distillation and the student model's fine-tuning. The three processes all use the InfoNCE as the objective functions.
The consistency can bring further performance improvement to the student model, just as prompt learning does.
Moreover, in the distillation process, the InfoNCE loss brings two enforcements to the student model's sentence embeddings: (1) be close to that learned by the teacher;  (2)  be far away from the sentence embeddings of other sentences learned by the teacher model. In contrast, previous KD methods using the MSE objective function only satisfy the first enforcement.


In our experiments, we use the well-trained sentence embedding model SimCSE-RoBERTa$_{large}$ (330M) as the teacher, and small Transformer-based models as the students, using different sizes of parameters (110M/52M/14M).
Evaluations on 7 STS benchmark datasets well demonstrate that our proposed DistilCSE framework and CKD method can compress the large sentence embedding model with little or even no performance drop.
Impressively, our 110M student model can slightly outperform the latest SOTA model, i.e., Sentence-T5(11B), with only \textbf{1\%} parameters and 0.25\% unlabeled data.

We summarize our contributions as follows:
\begin{itemize}[itemsep=0pt,topsep=1pt,parsep=1pt]
\item
We propose an efficient knowledge distillation framework for contrastive sentence embedding, called \textbf{DistilCSE}, by combining unsupervised knowledge distillation with supervised contrastive learning.
\item
We further propose Contrastive Knowledge Distillation (CKD) to achieve better distillation effect by enhancing the training objective consistency among teacher model training, knowledge distillation and student model fine-tuning.
\item
Experimental results on STS benchmarks well demonstrate the effectiveness of the proposed DistilCSE framework and CKD method. 
\end{itemize}

\section{Related Work}
\paragraph{Knowledge Distillation} 

Knowledge Distillation \cite{hinton2015distilling} is a commonly used model compression technique, and the knowledge distillation methods for pre-trained models have been extensively studied.
\cite{sanh2019distilbert, sun2019patient, jiao2019tinybert} proposes to distill the logits in the prediction layer or \textit{(and)} hidden representations in the intermediate layers from the teacher model into a student model. 
\cite{aguilar2020knowledge} formulates two ways to distill the internal knowledge to improve the student model's generalization capabilities. 
\cite{sun2020mobilebert} trains a specially designed teacher model, and transfers to a task-agnostic student model.
\cite{pan2020meta} proposes to build a meta-teacher model that captures transferable knowledge across domains.
The most related work to ours is \cite{sun2020contrastive}, which proposes to distill knowledge through intermediate layers of the teacher model via a contrastive objective function. 
Although \cite{sun2020contrastive} applies contrastive learning in knowledge distillation, it is quite different from our method. 
Like many previous KD methods, \cite{sun2020contrastive} is only optimized for classification tasks and ignores sentence embedding distillation, while we focus on effective contrastive sentence embedding distillation.
Furthermore, \cite{sun2020contrastive} applies contrastive distillation on the intermediate representations as an auxiliary loss to the original KD loss, while our method applies a contrastive learning loss on the output representations to substitute the original KD loss. This substitution further brings consistency of the objective functions to the whole process, which improves the student model.

\paragraph{Contrastive Sentence Embedding} 
Contrastive learning has emerged as a promising trend in learning sentence embeddings by pulling similar sentences closer and pushing away different sentences in the embedding space.
\cite{fang2020cert, giorgi2020declutr, wu2020clear, yan2021consert, gao2021simcse} propose to employ multiple augmentation strategies to learn noise-invariant unsupervised sentence embeddings while solving the collapse issue of BERT-derived sentence embeddings.
\cite{gao2021simcse} also incorporates labeled NLI sentence pairs in supervised contrastive learning and achieves remarkable performance.
The most related one is \cite{ni2021sentence} which achieves state-of-the-art performance on STS tasks with as high as 11B model parameters.
\cite{ni2021sentence} holds a multi-stage training framework similar to our method, which contrastively trains firstly on 2 Billion question-answers pairs and then on the NLI dataset.
Different from \cite{ni2021sentence}, we do not train a contrastive sentence embedding model from scratch. Instead, we focus on effective knowledge distillation for contrastive sentence embeddings with little or no performance decrease. 
Moreover, the unlabeled data used in our method is much smaller, only 0.25\% of \cite{ni2021sentence}.
\section{DistilCSE Framwork}
As illustrated in Figure \ref{DistilCSE}, the proposed DistilCSE framework consists of two stages. The stage 1  which will be introduced in section \ref{sec:stage1} and the stage 2  will be introduced in section \ref{sec:stage2}. 

\subsection{Knowledge Distillation on Unlabeled Data}
\label{sec:stage1}
In the proposed DistilCSE framework, the knowledge distillation stage follows the well-known teacher-student structure. It aims to transfer the capability of the large teacher model to the small student model. 
The teacher model is a parameter-fixed encoder with $T$ layers of transformer blocks trained with contrastive learning on labeled training data, e.g., NLI.
The student model is a to-be-learned encoder with $S$ ($< T$) layers transformer blocks, whose parameters can be initialized with pre-trained models like BERT.
To make the knowledge transfer more adequate, knowledge distillation is conducted on a large amount of unlabeled data in DistilCSE, rather than on the limited labeled training data used to train the teacher model.
The student model is trained to imitate the behavior of the teacher model.
Given a mini-batch with $N$ sentences $X=\{x_1, x_2, ..., x_N\}$. For each sentence $x_i \in X$, the teacher model and the student model will encode it into $h_i^T$ and $h_i^S$, respectively, as follows:
\begin{equation}
\begin{aligned}
& h_i^T = Teacher(x_i)*\\
& h_i^S = Student(x_i)
\end{aligned}
\end{equation}
, where $*$ means the parameters of the teacher model is fixed.

\paragraph{Contrastive Knowledge Distillation} 

Previously commonly used KD methods use MSE loss to force $h_i^S$ to be close to $h_i^T$ for each sentence $x_i$. MSE loss measures the difference between $h_i^S$ and $h_i^T$ with $L2$-norm for each $x_i$ in a mini-batch with $N$ sentences as follows:
\begin{equation}
\mathcal{L}_{\mathrm{KD}}=\sum_{i=1}^{N} MSE\left(h_i^S, h_i^T\right)
\end{equation}
In the case that the dimension of $h_i^T$ is different from that of $h_i^S$, a learnable linear projection matrix $M$ is needed to adjust the dimension of $h_i^S$ to be the same as $h_i^T$. Then the loss function above can be redefined as follows.
\begin{equation}
\label{eq:kdm}
\mathcal{L}_{\mathrm{KD}}=\sum_{i=1}^{N} MSE\left(h_i^S M, h_i^T\right)
\end{equation}

Our newly proposed contrastive knowledge distillation (CKD) method uses InfoNCE loss to enforce $h_i^S$ to be close to $h_i^T$ and meanwhile enforce $h_i^S$ to be away from $h_j^T$ corresponding to any other sentence $x_j$.
Given a sentence $x_i$ in a mini-batch, $h_i^S$ and $h_i^T$ form a positive pair, and meanwhile $h_i^S$ and $h_j^T$ form a negative pair, where $h_j^T$ is the sentence embedding of any other sentence $x_j$ within the mini-batch. Then we leverage the widely-used contrastive learning loss, i.e., the InfoNCE loss, to encourage $h_i^S$ to be close to $h_i^T$ and meanwhile away from $h_j^T$:
\begin{equation}
\mathcal{L}_{\mathrm{CKD}}=-\log \frac{e^{f\left(\mathbf{h}_{i}^{S}, \mathbf{h}_{i}^{T}\right) / \tau}}{\sum\limits_{j=1}^{N}\left(e^{f\left(\mathbf{h}_{i}^{S}, \mathbf{h}_{j}^{T}\right) / \tau}\right)}
\end{equation}
where $f\left(u, v\right)$ is the cosine similarity between $u$ and $v$, $\tau$ is a temperature hyperparameter.

The \textbf{memory bank mechanism} is widely adopted in contrastive learning\cite{he2020momentum,chen2020improved}. It allows reusing the encoded sentence embeddings from the immediate preceding mini-batches by maintaining a fixed size queue, which can enlarge the size of negative pairs for contrastive learning and thus bring performance gain. 
Here, we also incorporate the memory bank mechanism in the proposed CKD to allow the output embeddings of the student model to be compared with more output embeddings of the teacher model, without increasing the batch size. 
Specifically, we construct a memory bank queue for the output embeddings of the teacher model from consequent mini-batches. 
And the embeddings in the memory bank queue will be progressively replaced. When the sentence embeddings of the teacher model for the current mini-batch are enqueued, the ``oldest" ones in the queue are removed if the queue is full.
With the memory bank queue, the InfoNCE loss is further modified as follows:
\begin{equation}
\label{eq:CKD2}
\begin{aligned}
  &\mathcal{L}_{\mathrm{CKD}} = \\
  &-\log \frac{e^{f\left(\mathbf{h}_{i}^{S}, \mathbf{h}_{i}^{T}\right) / \tau}}{\sum\limits_{j=1}^{N}\left(e^{f\left(\mathbf{h}_{i}^{S}, \mathbf{h}_{j}^{T}\right) / \tau}\right) + \sum\limits_{q=1}^{Q}\left(e^{f\left(\mathbf{h}_{i}^{S}, \mathbf{h}_{q}^{T}\right) / \tau}\right)}
\end{aligned}
\end{equation}
where $h_{q}^{T}$ denotes a sentence embedding of the teacher model in the memory bank queue with a size of $Q$. 
Similarly, in the case that the dimension of $h_i^T$ is different from that of $h_i^S$, Equation \ref{eq:CKD2} is redefined as follows:
\begin{equation}
\label{eq:ckdm}
\begin{aligned}
  &\mathcal{L}_{\mathrm{CKD}} = \\
  &-\log \frac{e^{f\left(\mathbf{h}_{i}^{S} M, \mathbf{h}_{i}^{T}\right) / \tau}}{\sum\limits_{j=1}^{N}\left(e^{f\left(\mathbf{h}_{i}^{S} M, \mathbf{h}_{j}^{T}\right) / \tau}\right) + \sum\limits_{q=1}^{Q}\left(e^{f\left(\mathbf{h}_{i}^{S} M, \mathbf{h}_{q}^{T}\right) / \tau}\right)}
\end{aligned}
\end{equation}

\subsection{Student Finetuning with Contrastive Learning on Labeled Data}
\label{sec:stage2}
To alleviate the potential domain bias brought by a large amount of unlabeled data, we further conduct student model finetuning with contrastive learning on labeled data. The labeled data is the same data used to train the teacher model. The finetuning process enables the student model to fit the textual similarity measurement better.

Suppose that the original training data consists of tuples $\left(x_{i}, x_{i}^{+}, x_{i}^{-}\right)$, where $x_{i}$ is a sentence, $x_{i}^{+}$ is a similar sentence to $x_{i}$, and $x_{i}^{-}$ is a dissimilar one. 
We conduct contrastive learning to finetune the student model on the training data, using the InfoNCE loss as follows:
\begin{equation}
\label{eq:studentfinetune}
\mathcal{L}_{\mathrm{CL}}=-\log \frac{e^{f\left(\mathbf{h}_{i}^{S}, \mathbf{h}_{i}^{S+}\right) / \tau}}{\sum\limits_{j=1}^{N}\left(e^{f\left(\mathbf{h}_{i}^{S}, \mathbf{h}_{j}^{S+}\right) / \tau}+e^{f\left(\mathbf{h}_{i}^{S}, \mathbf{h}_{j}^{S-}\right) / \tau}\right)}
\end{equation}
where $N$ is the size of a mini-batch of sentences, $h_i^S$ and $h^{S+}_i$ denote the sentence embeddings of $x_i$ and $x^+_i$ output by the student model, respectively. 

After being finetuned with contrastive learning, the small student model can then be applied to real-world applications with generally much lower computational costs and little or even no performance decrease, as demonstrated by experiments below.

\section{Experiment}
In this section, we first introduce the setup and implementation details of our experiments, and then we analyze our experimental results.

\subsection{Experiment Setup}
\paragraph{Datasets} 
We construct an unlabeled dataset with 5M high-quality english sentences from open-source news, termed News-5m, for the knowledge distillation stage of the proposed DistilCSE framework.
And in the student model finetuning stage, we directly leverage the labeled NLI dataset, which is also the dataset used to train the large teacher model. Specifically, the NLI dataset consists of 275K sentence pairs, each being either an entailment hypothesis or a contradiction hypothesis for a premise (i.e., sentence). Following \cite{gao2021simcse}, we use the entailment pairs as positives and contradiction pairs as negatives to build the needed tuples for Eq \ref{eq:studentfinetune}.

\paragraph{Baselines} 
We compare with the latest SOTA models, i.e., 110M/330M/3B/11B Sentence-T5 \cite{ni2021sentence}, and 330M/110M/52M/14M SimCSE \cite{gao2021simcse}.
Note that Sentence-T5 explores a variety of experimental settings for each size of parameters, and we choose the best results for comparison.
The results of baseline models are reported from the corresponding papers, except for 52M and 14M SimCSE, which have no reported results or published models Thus we train 52M and 14M SimCSE by ourselves with the officially released code.
For the knowledge distillation method, we compared our proposed CKD with KD \cite{hinton2015distilling}, PKD \cite{sun2019patient} and Meta-KD \cite{pan2020meta}.  Supplementary note, according to the original paper of \cite{sun2020contrastive}, ``Our finetuning framework is designed for classification, we only exclude the STS-B dataset", so \cite{sun2020contrastive}'s method is not suitable for sentence embedding distillation and we cannot directly compare with it on the STS benchmarks.

\paragraph{Evaluation} We evaluate all methods on 7 widely used STS benchmarks, i.e., STS 2012–2016 \citep{agirre2012semeval,agirre2013sem,agirre2014semeval,agirre2015semeval,agirre2016semeval} and STS-B \cite{cer2017semeval}. to measure the semantic similarity of any two sentences with the cosine similarity between the corresponding sentence embeddings. 
After deriving the semantic similarities of all sentence pairs in the test set, we follow \cite{gao2021simcse} to use Spearman correlation\footnote{\url{https://en.wikipedia.org/wiki/Spearman\%27s_rank_correlation_coefficient}} to measure the correlation between the ranks of predicted similarities and that of the ground-truth similarities. 
Specially, we utilize the public SentEval toolkit\footnote{\url{https://github.com/facebookresearch/SentEval}} to evaluate the models on the dev set of STS-B to search for better settings of the hyper-parameters. Then the best-performing checkpoint is evaluated on the STS test sets.

\subsection{Training Details}
\begin{table}[!t]
\centering
\small
\begin{tabular}{cccc}
\toprule 
\textbf{Model} & \textbf{\#layers} & \textbf{\#embed} & \textbf{\#params} \\
 \hline
Teacher & 24 & 1024 & 330M \\
\hline
BERT$_{base}$ & 12 & 768 & 110M \\
Tiny-L6 & 6 & 768 & 52M \\
Tiny-L4 & 4 & 312 & 14M \\
\bottomrule
\end{tabular}
\caption{Model sizes of the teacher model and different student models. \#layers denotes model layers, \#embed denotes embedding size, \#params denotes parameter size.}
\label{model_statistics}
\end{table}

\begin{table*}[!t]
\centering
\small
\setlength{\tabcolsep}{1.4mm}{\begin{tabular}{lcccccccccl}
\toprule 
\textbf{\#Params} & \textbf{Stage} & \textbf{Model} & \textbf{STS12} & \textbf{STS13} & \textbf{STS14} & \textbf{STS15} & \textbf{STS16} & \textbf{STS-B} & \textbf{SICK-R} & \textbf{Avg.} \\
\midrule 
\midrule 
11B & - & Sentence-T5$\clubsuit$ & \textbf{80.11} & 88.78 & \textbf{84.33} & 88.36 & \textbf{85.55} & 86.82 & 80.60 & 84.94 \\
\midrule 
3B & - & Sentence-T5$\clubsuit$  & 79.02 & 88.80 & 84.33 & \textbf{88.89} & 85.31 & 86.25 & 79.51 & 84.59 \\ 
\midrule 
\multirow{2}{*}{330M} & - & Sentence-T5$\clubsuit$  & 79.10 & 87.32 & 83.17 & 88.27 & 84.36 & 86.73 & 79.84 & 84.11 \\
 & - & \underline{SimCSE-RBT$_{large}$ $\clubsuit$}  & \underline{77.46} & \underline{87.27} & \underline{82.36} & \underline{86.66} & \underline{83.93} & \underline{86.70} & \underline{\textbf{81.95}} & \underline{83.76} \\
\midrule 
\multirow{12}{*}{110M} & - & Sentence-T5$\clubsuit$  & 78.05 & 85.84 & 82.19 & 87.46 & 84.03 & 86.04 & 79.75 & 83.34 \\
 & - & SimCSE-RBT$_{base}$ $\clubsuit$ & 76.53 & 85.21 & 80.95 & 86.03 & 82.57 & 85.83 & 80.50 & 82.52 \\
 & - & SimCSE-BERT$_{base}$ $\clubsuit$ & 75.30 & 84.67 & 80.19 & 85.40 & 80.82 & 84.25 & 80.39 & 81.57 \\
 \cline{2-11} \\
 & \multirow{4}{*}{\textbf{stage1}} & KD-BERT$_{base}$ & 75.60 & 86.75 & 81.31 & 86.51 & 83.63 & 86.01 & 81.56 & 83.05 \\
 & & PKD-BERT$_{base}$ &  76.27 & 86.90  & 81.44 & 86.35  & 83.80 & 85.92  & 81.15 & 83.12  \\
 & & MetaKD-BERT$_{base}$ & 75.16  & 86.41  & 81.24 & 86.55  & 83.41 & 85.87  & 81.32 &  82.85 \\
 & & CKD-BERT$_{base}$  & 76.48 & 86.94 & 82.42 & 87.37 & 83.65 & 86.27 & 81.03 & 83.45 \\ 
  \cline{2-11} \\
 & \multirow{4}{*}{\textbf{stage2}} & KD-BERT$_{base}$ & 78.57 & 88.32 & 83.52 & 87.85 & 84.56 & 87.60 & 81.55 & 84.57 \\
 & & PKD-BERT$_{base}$ &  78.54 & 88.14  & 82.98 & 87.56  & 84.40 & 87.28  & 81.49 & 84.34  \\
 & & MetaKD-BERT$_{base}$ & 78.22  & 87.69  & 83.16 & 87.92  & 84.55 & 87.30  & 81.25 & 84.30  \\
 & & \textbf{CKD-BERT$_{base}$} & 79.51 & \textbf{88.85} & 84.10 & 88.47 & 85.06 & \textbf{87.97} & 81.34 & \textbf{85.04} \\
\midrule
\multirow{10}{*}{52M} & -  & SimCSE-Tiny-L6$\spadesuit$ & 75.66	& 83.49	& 79.82	& 85.14	& 80.41	& 83.08	& 80.00 & 81.09 \\
 \cline{2-11} \\
& \multirow{4}{*}{\textbf{stage1}} & KD-Tiny-L6 & 75.19 & 86.26 & 80.64 & 86.60 & 82.52 & 84.81 & 80.04 & 82.29 \\
& & PKD-Tiny-L6  & 74.80  & 86.43  & 80.94 & 86.38  & 83.19 & 85.78  & 81.13 & 82.66  \\
& & MetaKD-Tiny-L6  & 75.47  & 86.31  & 80.83 & 86.11  & 83.33 & 85.53  & 80.94 & 82.65  \\
&  & CKD-Tiny-L6 & 75.83 & 86.88 & 82.12 & 87.61 & 83.51 & 85.97 & 80.26 & 83.17 \\ 
\cline{2-11} \\
 & \multirow{4}{*}{\textbf{stage2}} & KD-Tiny-L6 & 77.97 & 87.32 & 82.89 & 87.56 & 83.85 & 86.46 & 80.89 & 83.85 \\
&  & PKD-Tiny-L6  & 78.52  & 87.96  & 83.33 & 87.67  & 84.83 & 87.19  & 80.77 & 84.32  \\
&  & MetaKD-Tiny-L6  & 78.13  & 87.96  & 83.15 & 87.61  & 84.43 & 87.08  & 81.07 & 84.20  \\
&  & \textbf{CKD-Tiny-L6} & 78.20 & 88.21 & 83.75 & 88.50 & 84.61 & 87.53 & 81.29 & 84.58 \\
\midrule
\multirow{9}{*}{14M} & - &  SimCSE-Tiny-L4$\spadesuit$ & 74.90 & 78.07 & 73.56 & 81.51 & 77.24 & 77.78 & 77.30 & 77.19 \\ 
 \cline{2-11} \\
& \multirow{4}{*}{\textbf{stage1}} & KD-Tiny-L4 & 74.41 & 83.84 & 78.89 & 84.71 & 80.45 & 82.78 & 77.93 & 80.43 \\
& & PKD-Tiny-L4  & 74.35  & 83.38  & 77.56 & 84.26  & 79.14 & 82.56  & 77.79 & 79.86  \\
& & MetaKD-Tiny-L4  & 73.4  & 84.60  & 79.13 & 84.65  & 80.45 & 83.10  & 78.36 & 80.53  \\
& & CKD-Tiny-L4 & 74.27 & 84.71 & 80.19 & 85.41 & 81.94 & 83.63 & 77.94 & 81.16 \\ 
\cline{2-11} \\
& \multirow{4}{*}{\textbf{stage2}} & KD-Tiny-L4 & 76.58 & 85.40 & 81.76 & 86.72 & 82.71 & 84.87 & 79.89 & 82.56 \\
& & PKD-Tiny-L4  & 76.64  & 85.74  & 81.33 & 86.47  & 82.39 & 84.94  & 80.03 & 82.51  \\
& & MetaKD-Tiny-L4  & 76.71  & 86.03  & 81.67 & 86.80  & 82.61 & 84.81  & 80.16 & 82.68  \\
& & \textbf{CKD-Tiny-L4} & 77.24 & 85.50 & 81.94 & 87.10 & 82.97 & 85.16 & 80.00 & 82.84 \\
\bottomrule
\end{tabular}}
\caption{Sentence embedding performance on 7 semantic textual similarity (STS) test sets, in terms of Spearman’s correlation. ``\textbf{stage1}'' and ``\textbf{stage2}'' are the two stages of DistilCSE, as shown in Figure \ref{DistilCSE}.
$\clubsuit$ : results from \cite{reimers2019sentence, gao2021simcse}. $\spadesuit$: Small SimCSE models trained by ourselves with the code and data from \cite{gao2021simcse}. Results with \underline{underline} is the teacher model's performance. ``RBT'' is short for RoBERTa.
}
\label{table_test_best}
\end{table*}

\paragraph{Model Settings} 
For the teacher model, we choose the 330M pre-trained checkpoint of SimCSE-RoBERTa$_{large}$, which composes of 24 layers of transformer block. An MLP layer is added on top of the [CLS] representation to get the sentence embedding, and the dimension of sentence embeddings is 1024. During training, the parameters of the teacher model are fixed and will not be updated.
For the student model, we have three different settings of the parameter size, i.e., 110M/52M/14M. They are in similar network structures to the teacher model, except that the number of layers and the dimensions of sentence embeddings are smaller. The 110M models are initialized from the pre-trained BERT$_{base}$, and the 52M and 42M models are initialized from the pre-trained TinyBERT.
We list the model information of the teacher model and the different student models in Table \ref{model_statistics}.
Note that the dimension of sentence embeddings output by each student model is different from that of the teacher model, and thus for the knowledge distillation state in the proposed DistilCSE framework, we need to add a layer of linear projection upon the embeddings output by the student model to map them to 1024-dimensional ones, as denoted by Eq. \ref{eq:kdm} and Eq. \ref{eq:ckdm}.

\paragraph{Optimization Settings}
In the knowledge distillation stage on large unlabeled data, we train our student models for 20 epochs, using the Adam \cite{kingma2014adam} optimizer with a batch size of $512$. Learning rate is set as $2e^{-4}$ for 110M model, and $3e^{-4}$ for 52M and 14M models. 
Particularly for the proposed CKD, we follow \cite{ALBEF} to use a memory bank queue with the size being $65536$.
Following \cite{gao2021simcse}, we evaluate the performance of each student model every 125 training steps on the dev set of STS-B.
The training process uses the early-stop strategy with the patience being $3$, which stops if the performance on the dev set is not updated for 3 consecutive epochs.

In the student finetuning stage, 
we load each student model from the best performing checkpoint in the knowledge distillation stage and train it for 5 epochs using the Adam optimizer, with a batch size of $128$. Learning rate is set as $1e^{-5}$ for 52M model, and $5e^{-5}$ for 110M and 14M models. 
Each student model is still evaluated every 125 training steps on the dev set of STS-B, and the best checkpoint is used for the final evaluation on test sets.

\subsection{Experiment Results}

In Table \ref{table_test_best}, we report the performance of the student models trained using the DistilCSE framework in two stages.
In all settings of parameter size, student models trained through the proposed DistilCSE framework outperform their counterparts in Figure \ref{probe_exp}. This fully illustrates the usefulness of the DistilCSE framework. For those KD methods that are not suitable for the contrastive sentence embedding distillation task, DistilCSE enables them to achieve as good performance as on other tasks.
Furthermore, our proposed CKD consistently achieves better results than the previous KD/PKD/MetaKD methods in both stages, which verifies that objective function consistency in different processes is beneficial to improving the performance of the student models.
Specifically, the objective function consistency of teacher constative learning and contrastive knowledge distillation enables CKD to outperform other KD methods at stage 1. The objective function consistency of contrastive knowledge distillation and student contrastive finetuning allows CKD to achieve the best at stage 2.

\paragraph{110M Student Models:} All DistilCSE-*-BERT$_{base}$ student models achieve better performance than the teacher model, i.e., SimCSE-RoBERTa$_{large}$, with 1/3 parameters size. And amazingly, DistilCSE-CKD-BERT$_{base}$ can slightly outperform the latest SOTA 11B Sentence-T5, with only 1\% parameters and 0.25\% unlabeled data. Meanwhile, DistilCSE-KD-BERT$_{base}$ can achieve comparable performance to the 3B Sentence-T5. 

\paragraph{52M Student Models:} All DistilCSE-*-Tiny-L6 student models outperform the teacher model, i.e., SimCSE-RoBERTa$_{large}$, with 1/6 parameters size. And DistilCSE-CKD-Tiny-L6 achieves comparable performance to the 3B Sentence-T5, with only 1/60 parameters.

\paragraph{14M Student Models:} All DistilCSE-*-Tiny-L4 and DistilCSE-*-Tiny-L4 achieve comparable or higher performance to the SimCSE-BERT$_{base}$ model, with only 1/8 parameters.\\

The analyses above show that the small student models derived from the proposed DistilCSE framework can even outperform the large teacher model. We argue that it can be attributed to both following potential reasons. At stage 1, knowledge distillation on large unlabeled data can not only adequately transfer the model capability of the teacher model to the student model, but also allow the student model view more data, which is somehow like a semi-supervised setting. At stage 2, the student model has been initialized close to the local optimum of the teacher model, and further fine-tuning process through contrastive learning may bring it to another better local optimum, resulting in a performance boost.

\section{Further Analyses}
In this section, we conduct some analyses on the DistilCSE framework. Following \cite{gao2021simcse}, all results are evaluated on the development set of STS-B unless otherwise specified. The DistilCSE-* models used below are all from stage 2.

\begin{figure}[!tbp]
\centering
\includegraphics[width=0.5\textwidth,height=0.2\textheight]{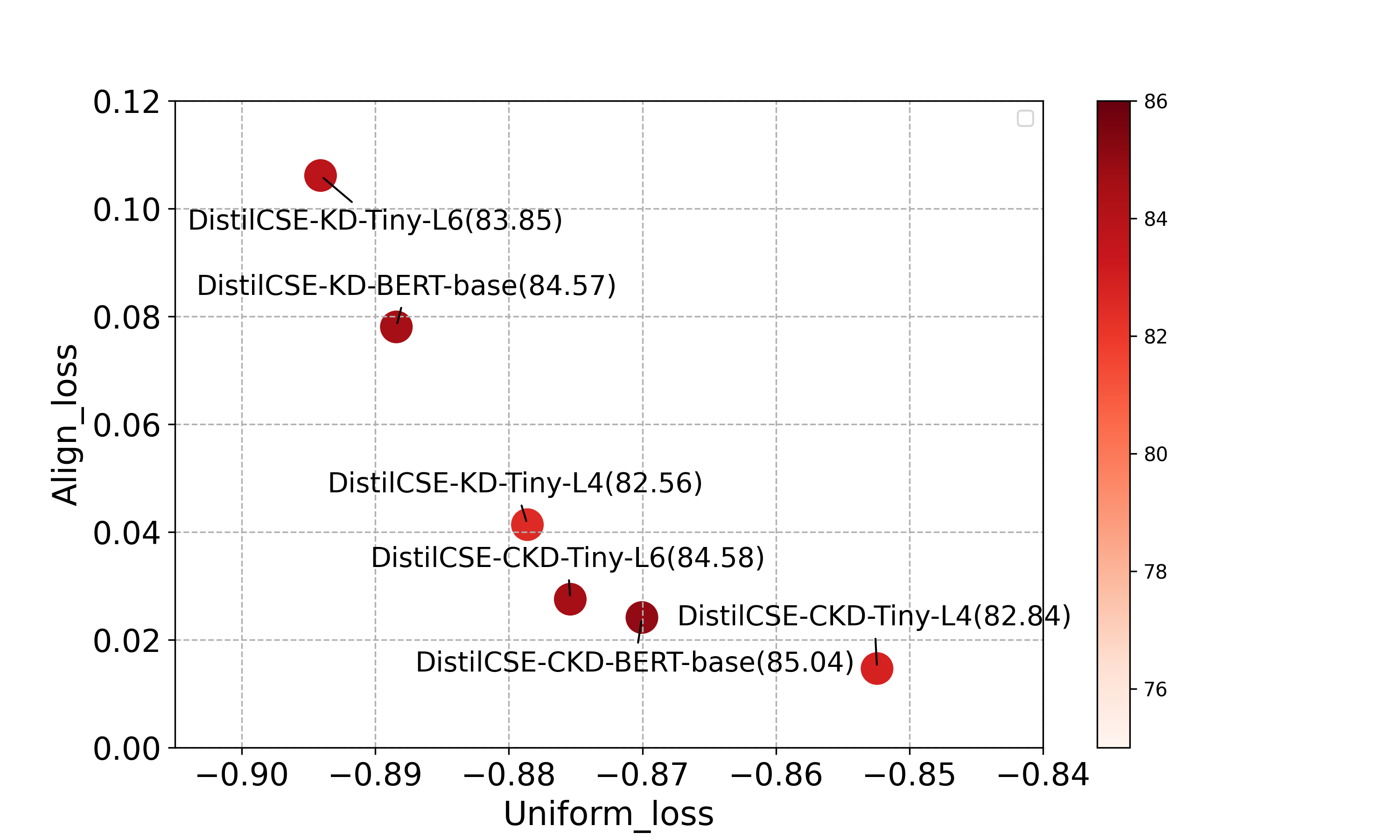}
\caption{$\ell_{\text {align }}-\ell_{\text {uniform }}$ plot for student models. All models are trained through the DistilCSE framework. For ease of presentation, we use abbreviations in the figure.}
\label{align_uniform}
\end{figure}

\subsection{Uniformity and Alignment}
We investigate the alignment and uniformity of models of different parameter sizes and knowledge distillation methods in the DistilCSE framework. Following \cite{wang2020understanding}, we compute the alignment loss and uniformity loss to measure the quality of the learned sentence embeddings, which are defined as follows.
\begin{equation}
\begin{gathered}
\mathcal{L}_{\text {align }}=-\underset{v, v^{+} \sim p_{\text {pos }}}{\mathbb{E}}\left\|f(v)-f\left(v^{+}\right)\right\| \\
\mathcal{L}_{\text {uniform }}=\log \underset{v, w \stackrel{\mathrm{i}, i, d}{\sim} p_{\text {data }}}{\mathbb{E}} \quad e^{-2\|f(v)-f(w)\|}
\end{gathered}
\end{equation}
where $p_{pos}$ denotes all positive pairs of similar sentences, and $p_{data}$ is the data distribution. 
$L_{align}$ is the expected distance between the embeddings of the two sentences in a positive pair, and $L_{uniform}$ denotes the uniformity of the embedding distribution. For both $L_{align}$ and $L_{uniform}$, a lower value indicates better performance. 

As shown in Figure \ref{align_uniform}, DistilCSE-CKD models can achieve better alignment, while DistilCSE-KD models can have better uniformity. Meanwhile, among student models with different parameter sizes, the 6-layer ones can achieve better uniformity, the 4-layer ones can achieve better alignment, and the 12-layer ones can reach a balance in good uniformity and alignment.

\subsection{Effects of the Scales of Unlabeled Data for KD}

\begin{figure}[!tbp]
\centering
\includegraphics[width=0.5\textwidth,height=0.2\textheight]{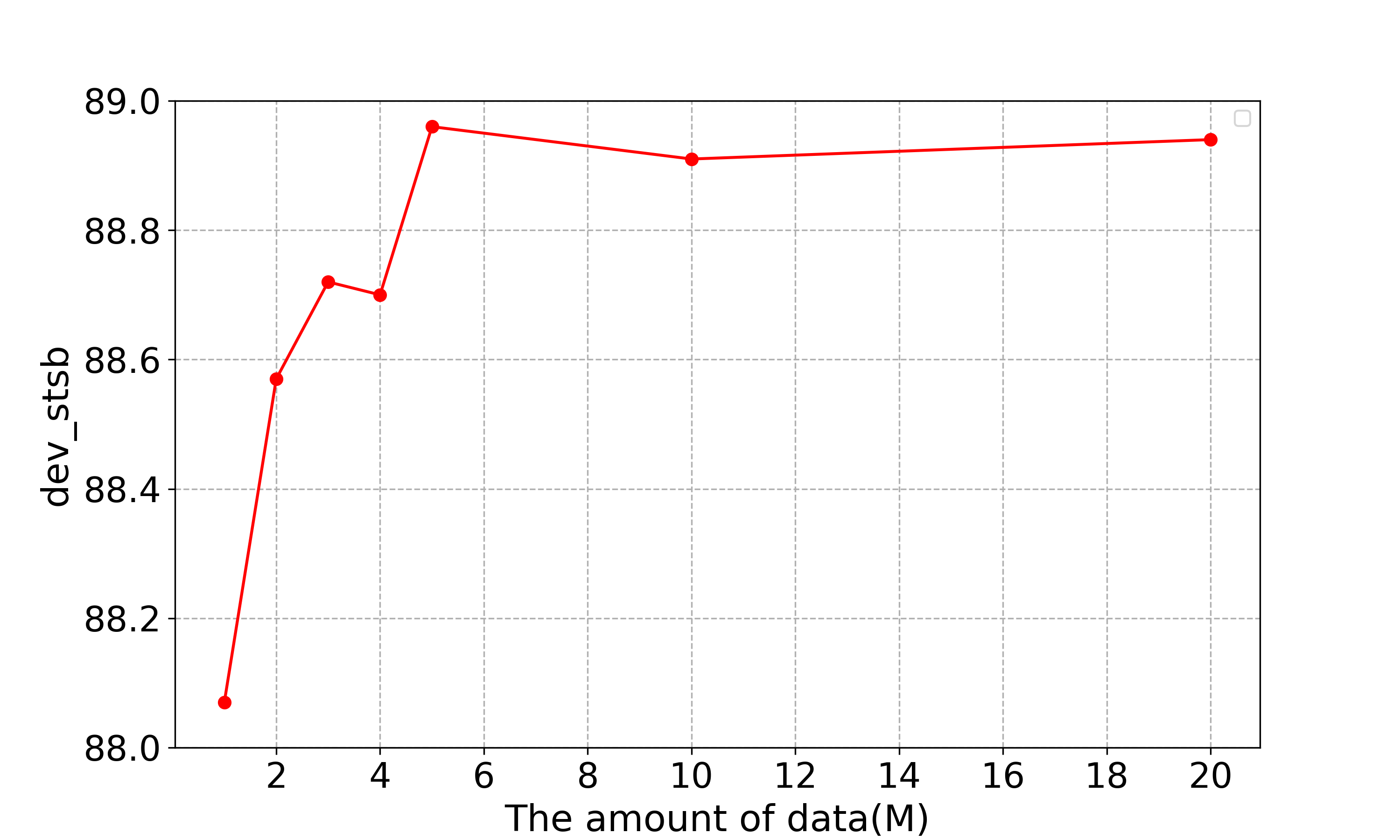}
\caption{Effects of different scales of the unlabeled data in the KD stage on DistilCSE-CKD-BERT$_{base}$, evaluated on STS-B development set, in terms of Spearman’s correlation.}
\label{data_scale}
\end{figure}

We investigate the effects of different scales of the unlabeled
data in the KD stage on DistilCSE-CKD-BERT$_{base}$ model. 
We increase the dataset scale from 1M to 20M gradually and plot the best results on STS-B development set in Figure \ref{data_scale}.
As the scale of unlabeled data for the CKD stage increases, the performance of DistilCSE-CKD-BERT$_{base}$ will firstly increase then tend to converge.
Specifically, when the amount of unlabeled data increases from 1M to 5M, Spearman's correlation on the STS-B development set keeps increasing, which indicates that the amount of unlabeled data is still critical to the proposed DistilCSE framework.
Yet when the amount of unlabeled data is larger than 5M, the Spearman's correlation on the STS-B development set tends to converge and varies between 88.9 and 89.0. Compared with the 2 billion question-answer pairs used in Sentence-T5 (11B), the proposed DistilCSE framework is kind of data-efficient during the KD stage and thus very large unlabeled data is not needed.


\subsection{Initializing Student Model with SimCSE}
We further explore whether initializing the student models with the parameters of a well-trained SimCSE models can bring further improvement, as SimCSE can yield significantly superior performance than BERT for STS.

In table \ref{student_use_simcse}, we report the evaluation results of student models initialized from either the well-trained SimCSE or the general pre-trained models (i.e., BERT$_{base}$, and TinyBERT) in different settings of parameter sizes, i.e., 110M/52M/14M.
It can be seen that initialization with SimCSE achieves comparable performance to that of initialization with general pre-trained models, which reflects the robustness of the proposed DistilCSE framework in some sense.

\begin{table}[!tbp]
\centering
\small
\begin{tabular}{ccc}
\hline 
\textbf{\#Params}& \textbf{Initialization}& \textbf{dev\_stsb} \\
\hline
\multirow{2}{*}{110M} & BERT$_{base}$ &  88.96 \\
 & SimCSE &  \textbf{89.03}\\
\hline
\multirow{2}{*}{52M} & TinyBERT &  \textbf{88.74} \\
 & SimCSE &   88.70 \\
\hline
\multirow{2}{*}{14M} & TinyBERT &  87.10 \\
 & SimCSE &  \textbf{87.21} \\
\hline 
\end{tabular}
\caption{Evaluation results of student models initialized from the well-trained SimCSE or general pre-trained models (i.e., BERT$_{base}$ and TinyBERT) in different settings of parameter sizes. 
}
\label{student_use_simcse}
\end{table}

\section{Conclusion}
In this paper, we propose a two-stages framework termed \textbf{DistilCSE}, to compress large sentence embedding models with little or even no performance decrease.
We further propose a novel contrastive learning based knowledge distillation method termed Contrastive Knowledge Distillation (CKD), which can achieve better results than previous knowledge distillation methods in the proposed DistilCSE framework.
Experimental results on 7 STS benchmarks show that, the proposed DistilCSE and CKD are effective, and the learned student model can even slightly outperform the latest SOTA model Sentence-T5 (11B) with only 1\% parameters and 0.25\% unlabeled data.


\bibliography{anthology,custom}
\bibliographystyle{acl_natbib}
\newpage
\appendix
\section{Knowledge Distillation Experiment on the NLI Dataset}
we explore the effect of current state-of-the-art KD methods\cite{hinton2015distilling, sun2019patient,pan2020meta} on the contrastive sentence modelling task.
Specifically, we distill the well-trained SimCSE-RoBERTa$_{large}$ \cite{gao2021simcse} with 330 million parameters to a small student model with fewer parameters. As the teacher model is trained on the NLI dataset, we minimize the difference between the teacher's embedding and student's embedding for each sentence from the NLI dataset. 

We apply the KD process on the NLI data set under different parameter scales. 
Each student model is still evaluated every 125 training steps on the development set of STS-B, and the best checkpoint is used for the final evaluation on test sets \cite{gao2021simcse}.
The performances of student models on 7 semantic textual similarity (STS) test sets are shown in Table \ref{probe_exp}. 
The performance of the student models are substantially lower than the teacher model in the STS task evaluation. Thus, it is challenging to use only a single KD process to transfer the capability of the large teacher model adequately to the small student model, especially on the limited NLI training data. 

\begin{table*}[!htbp]
\centering
\setlength{\tabcolsep}{1mm}{\begin{tabular}{llcccccccl}
\toprule 
\textbf{\#Params} & \textbf{Model} & \textbf{STS12} & \textbf{STS13} & \textbf{STS14} & \textbf{STS15} & \textbf{STS16} & \textbf{STS-B} & \textbf{SICK-R} & \textbf{Avg.} \\
\midrule 
\midrule 
330M & SimCSE-RBT$_{large}$ $\clubsuit$  & 77.46 & 87.27 & 82.36 & 86.66 & 83.93 & 86.70 & 81.95 & 83.76 \\
\midrule 
\multirow{3}{*}{110M} & KD-BERT$_{base}$ & 74.92 & 85.50 & 80.04 & 85.10 & 82.50 & 84.92 & 80.70 & 81.95(-1.81) \\
 & PKD-BERT$_{base}$ &  75.70 & 85.78 & 80.31 & 85.54 & 82.84 & 84.87 & 80.58 & 82.23(-1.53)  \\
 & MetaKD-BERT$_{base}$ & 73.71 & 83.69 & 78.38 & 84.83 & 81.16 & 83.82 & 80.11 & 80.81(-2.95) \\
\midrule 
\multirow{3}{*}{52M} & KD-Tiny-L6 & 74.86 & 84.97 & 79.81 & 85.56 & 81.79 & 84.56 & 80.70 & 81.75(-2.01) \\
& PKD-Tiny-L6  & 74.19 & 84.71 & 79.50 & 85.28 & 82.09 & 84.97 & 80.74 & 81.64(-2.12)  \\
& MetaKD-Tiny-L6  & 75.47  & 86.31  & 80.83 & 86.11  & 83.33 & 85.53  & 80.94 & 82.65(-1.11)  \\
\midrule 
\multirow{3}{*}{14M} & KD-Tiny-L4 & 72.07 & 81.26 & 76.81 & 83.95 & 79.37 & 81.64 & 79.30 & 79.20(-4.56) \\
& PKD-Tiny-L4  & 72.21 & 79.73 & 75.65 & 83.04 & 79.33 & 81.67 & 79.47 & 78.73(-5.03)  \\
& MetaKD-Tiny-L4  & 72.04 & 79.73 & 74.61 & 83.47 & 77.08 & 79.52 & 77.92 & 77.77(-5.99)  \\
\bottomrule
\end{tabular}}
\caption{The embedding knowledge distillation experiment results on the NLI Dataset. $\clubsuit$ : teacher model. ``RBT'' is short for RoBERTa. Though these methods achieve very good results on other tasks, but they inevitably bring some drops in embedding effect.
}
\label{probe_exp}
\end{table*}

\section{Performance on Transfer Tasks}
Following \cite{gao2021simcse}, we further evaluate the performance of the proposed DistilCSE framework on transfer tasks, to see the transferability of the sentence embeddings output by the student models learned through DistilCSE. 
The transfer tasks include: MR (movie review) \cite{pang2005seeing}, CR (product review) \cite{hu2004mining}, SUBJ (subjectivity status) \cite{ pang2004sentimental} , MPQA (opinion-polarity) \cite{wiebe2005annotating}, SST-2 (binary sentiment analysis) \cite{socher2013recursive}, TREC (question-type classification) \cite{voorhees2000building} and MRPC (paraphrase detection) \cite{dolan2005automatically}. For more details, one can refer to SentEval\footnote{\url{https://github.com/facebookresearch/SentEval} }.

Following \cite{gao2021simcse}, we train a logistic regression classifier on the frozen sentence embeddings generated by different methods, and use the default configuration of SentEval for evaluation.
The evaluation results on the transfer tasks are reported in Table \ref{table_transfer}.
It can be seen that, the student models trained through DistilCSE-CKD consistently outperform the corresponding counterparts of the same parameter size trained on the NLI dataset. The results also validates the effectiveness of the proposed DistilCSE framework and CKD method. However, as \cite{gao2021simcse} argues, transfer tasks are not the major goal for sentence embeddings. Thus we leave it in appendix and still take the STS results for main comparison.

\begin{table*}[!t]
\centering
\setlength{\tabcolsep}{0.9mm}{\begin{tabular}{llcccccccl}
\toprule 
\textbf{\#Params} & \textbf{Model} & \textbf{MR} & \textbf{CR} & \textbf{SUBJ} & \textbf{MPQA} & \textbf{SST} & \textbf{TREC} & \textbf{MRPC} & \textbf{Avg.} \\
\midrule 
\midrule
110M & Sentence-T5$\clubsuit$ & 86.56 & 91.31 & 96.01 & 90.57 & 90.77 & 94.60 & 72.93 & 88.96 \\
 & SimCSE-RoBRTEa$_{base}$$\clubsuit$ & 84.92 & 92.00 & 94.11 & 89.82 & 91.27 & 88.80 & 75.65 & 88.08 \\
 & SimCSE-BRTE$_{base}$$\clubsuit$ & 82.68 & 88.88 & 94.52 & 89.82 & 88.41 & 87.60 & 76.12 & 86.86 \\
 &  DistilCSE-CKD-BERT$_{base}$ &  86.50 & 91.34 & 95.16 & 91.21 & 91.76 & 90.60 & 76.93 & \textbf{89.07} \\
 \midrule
 \multirow{3}{*}{52M} & SimCSE-Tiny-L6 & 81.96 & 88.93 & 94.30 & 89.84 & 86.66 & 88.20 & 75.25 & 86.45 \\
 & DistilCSE-CKD-Tiny-L6 &  85.61 & 90.97 & 94.78 & 91.43 & 90.99 & 90.60 & 76.70 & \textbf{88.73} \\
 \midrule
  \multirow{3}{*}{14M} & SimCSE-Tiny-L4 & 78.00 & 86.28 & 92.11 & 89.12 & 83.86 & 84.80 & 74.09 & 84.04 \\
 & DistilCSE-CKD-Tiny-L4 &  81.60 & 90.25 & 92.52 & 90.36 & 87.20 & 86.00 & 75.07 &\textbf{86.14} \\
\bottomrule
\end{tabular}}
\caption{Results on transfer tasks of different sentence embedding models, in terms of accuracy. $\clubsuit$ : results from \cite{gao2021simcse, ni2021sentence}. 
}
\label{table_transfer}
\end{table*}
\end{document}